\documentclass[letterpaper]{article} % DO NOT CHANGE THIS
\usepackage{aaai2027}  % DO NOT CHANGE THIS
\usepackage[hyphens]{url}  % DO NOT CHANGE THIS
\usepackage{graphicx} % DO NOT CHANGE THIS
\usepackage{natbib}  % DO NOT CHANGE THIS AND DO NOT ADD ANY OPTIONS TO IT
\usepackage{caption} % DO NOT CHANGE THIS AND DO NOT ADD ANY OPTIONS TO IT
\usepackage{algorithm}
\usepackage{algorithmic}

\usepackage{newfloat}
\usepackage{listings}
\DeclareCaptionStyle{ruled}{labelfont=normalfont,labelsep=colon,strut=off} % DO NOT CHANGE THIS
\floatstyle{ruled}
\newfloat{listing}{tb}{lst}{}
\floatname{listing}{Listing}

\usepackage{booktabs}

\usepackage{graphicx} % DO NOT CHANGE THIS
\usepackage{natbib}  % DO NOT CHANGE THIS AND DO NOT ADD ANY OPTIONS TO IT
\usepackage{caption} % DO NOT CHANGE THIS AND DO NOT ADD ANY OPTIONS TO IT
\usepackage[utf8]{inputenc} % allow utf-8 input
\usepackage[T1]{fontenc}    % use 8-bit T1 fonts
\usepackage{url}            % simple URL typesetting
\usepackage{booktabs}       % professional-quality tables
\usepackage{amsfonts}       % blackboard math symbols
\usepackage{nicefrac}       % compact symbols for 1/2, etc.
\usepackage{microtype}      % microtypography
\usepackage{xcolor}         % colors
\usepackage{amsmath}
\usepackage{graphicx}
\usepackage{subcaption}
\usepackage{tabularx}
\usepackage{algorithm}
\usepackage{algorithmic}
\usepackage{booktabs}
\usepackage{makecell}
\usepackage{multirow}
\usepackage{pifont}
\usepackage{enumitem}

\usepackage{newfloat}
\usepackage{listings}
\DeclareCaptionStyle{ruled}{labelfont=normalfont,labelsep=colon,strut=off} % DO NOT CHANGE THIS
\floatstyle{ruled}
\newfloat{listing}{tb}{lst}{}
\floatname{listing}{Listing}

\usepackage{booktabs}

\title{Identifying Informative Environments for Cognition Parameter Inference via Bayesian Experimental Design}

\author {
    Manisha Dubey ,
    Rimvydas Rubavicius ,
    N. Siddharth ,
    Subramanian Ramamoorthy 
}
\affiliations {
    School of Informatics, University of Edinburgh, United Kingdom\\
    mdubey@ed.ac.uk, rimvydas.rubavicius@ed.ac.uk, n.siddharth@ed.ac.uk, s.ramamoorthy@ed.ac.uk
}

\begin{document}

\maketitle

\begin{abstract}
Computational cognitive modeling seeks to infer latent cognitive mechanisms underlying observed behavior. Bayesian inverse planning provides a principled framework for such inference, but its success depends critically on the experimental environment. Existing approaches typically treat environments as fixed, leaving open the question of which cognitive experiments are most informative for cognition parameter inference. We formulate the design of cognitive planning experiments as a Bayesian Experimental Design (BED) problem, treating the experimental environment as the design variable. We establish an exact Monte Carlo BED benchmark and introduce an amortized Bayesian experimental design framework for efficient posterior inference and design evaluation. Experiments on the Mouselab-MDP process-tracing paradigm show that amortized BED closely matches the environment rankings of exact Monte Carlo BED while substantially reducing computational cost. We further show that no single environment is uniformly optimal across cognitive inference objectives, revealing trade-offs between expected information gain, posterior recoverability, and information efficiency. These results provide a principled framework for designing informative cognitive experiments for Bayesian parameter inference.
\end{abstract}

\section{Introduction}
Planning is a fundamental component of human cognition, yet the cognitive processes underlying sequential decision making are inherently latent and cannot be observed directly. Computational cognitive models increasingly rely on Bayesian inverse planning to infer latent cognitive mechanisms from observed behavior by fitting probabilistic generative models of planning. These mechanisms include planning horizon, working memory, attentional biases, and other resource-bounded cognitive processes~\cite{baker2009action,gelpi2025towards,zhang2025autotom,zhu2025capturing,banerjee2025estimating}. Recent advances show that large-scale behavioral datasets enable increasingly accurate computational models of human behavior~\cite{peterson2021using,binz2025foundation}. Game-based cognitive experiments further enable scalable behavioral data collection for studying increasingly complex planning behavior~\cite{allen2024using,fernandez2024expert,ying2025assessing}. At the same time, recent work shows that strategic reasoning is highly context dependent, varying systematically with properties of the decision environment rather than fixed cognitive parameters \cite{zhu2025capturing}. These findings suggest that the choice of experimental environment directly determines the behavioral evidence available for cognitive inference, motivating methods that actively design environments to maximize information about latent cognitive parameters.

Despite this growing evidence, existing inverse planning methods typically treat experimental environments as fixed or randomly sampled rather than optimizing them for cognitive inference. Some environments naturally expose latent cognitive differences, whereas others produce behavior that is intrinsically ambiguous. This raises a fundamental question: \textit{which experimental environments are maximally informative for recovering latent cognitive parameters?} Bayesian Experimental Design (BED) provides a principled framework for answering this question by selecting experiments that maximize the expected information gained about latent parameters before behavioral data are collected.

Applying BED to cognition parameter inference is computationally demanding because estimating the expected information gain of each candidate environment requires repeated Bayesian posterior inference over many simulated trajectory datasets, making exhaustive design optimization prohibitively expensive. Bayesian experimental design has previously been applied to optimize behavioral experiments for cognitive model discrimination and parameter estimation, primarily by selecting experimental conditions within fixed behavioral tasks \cite{an2014optimally, valentin2024designing}. In contrast, we treat the experimental planning environment itself as the design variable, enabling the automatic design of informative cognitive experiments for Bayesian inverse planning.

In this work, we combine Bayesian inverse planning with Bayesian Experimental Design to automatically identify informative sequential decision-making environments. We establish an exact Monte Carlo BED benchmark, develop an amortized surrogate for efficient posterior inference and design evaluation, and systematically analyze the properties that make environments informative through comparisons with heuristic and supervised baselines. Our key contributions are:
\begin{itemize}[noitemsep]
    \item We formulate the design of cognitive planning environments as a Bayesian Experimental Design problem over Bayesian inverse planning models, and establish an exact Monte Carlo BED benchmark for evaluating environment informativeness.
    
    \item We show that amortized Bayesian experimental design achieves environment rankings that closely agree with those produced by exact Monte Carlo BED while substantially reducing computational cost.
    
    \item We systematically evaluate exact Bayesian inference, Monte Carlo BED, and amortized BED on Mouselab-MDP, a widely used process-tracing paradigm for studying human planning across cognitive parameter inference tasks.
    
    \item We characterize informative environments through comparisons with structural heuristics and supervised baselines, revealing trade-offs between expected information gain, posterior recoverability, and information efficiency.
\end{itemize}

\section{Problem Setup}
We study the problem of identifying experimental environments that maximize the information available for inferring latent cognitive parameters from observed sequential decision-making behaviour. Let $\theta\in\Theta$ denote latent cognitive parameters that govern the agent's internal decision-making process and influence behavior only through the resulting action-selection policy. Different cognitive parameter settings induce different behavioural distributions under the same environment, enabling inference from observed trajectories. An environment $\xi\in\Xi$ is instantiated as a Markov Decision Process (MDP). Together with the latent cognitive parameters induces a distribution over behavioural trajectories $\mathcal D=\{\tau_i\}_{i=1}^N,\tau=(s_0,a_0,\ldots,s_T)$
according to $\mathcal D\sim p(\mathcal D\mid\theta,\xi)$. We use recoverability to denote the ability of an experimental environment to support accurate inference of latent cognitive parameters from observed behaviour. Following the Bayesian Experimental Design literature, we quantify recoverability using the expected information gain (EIG), which measures the expected reduction in posterior uncertainty over the latent parameters after observing behavioural trajectories as:
\begin{equation}
\label{eq:eig_entropy}
\mathrm{EIG}(\xi)
=
H(\theta)
-
\mathbb{E}_{p(\mathcal D\mid\xi)}
\!\left[
H(\theta\mid\mathcal D,\xi)
\right]
\end{equation}
Our objective is therefore to identify the environment that maximizes expected information gain
\begin{equation}
\label{eq:best_eig}
\xi^*
=
\arg\max_{\xi}
\mathrm{EIG}(\xi)
\end{equation}

Evaluating EIG requires repeated Bayesian posterior inference across many simulated datasets and candidate environments, making exact optimization computationally expensive. The remainder of the paper develops scalable approximations for efficient evaluation of the design objective.

\section{Methodology}
\label{sec:methodology}
\begin{figure}
    \centering
    \includegraphics[width=0.48\textwidth,keepaspectratio]{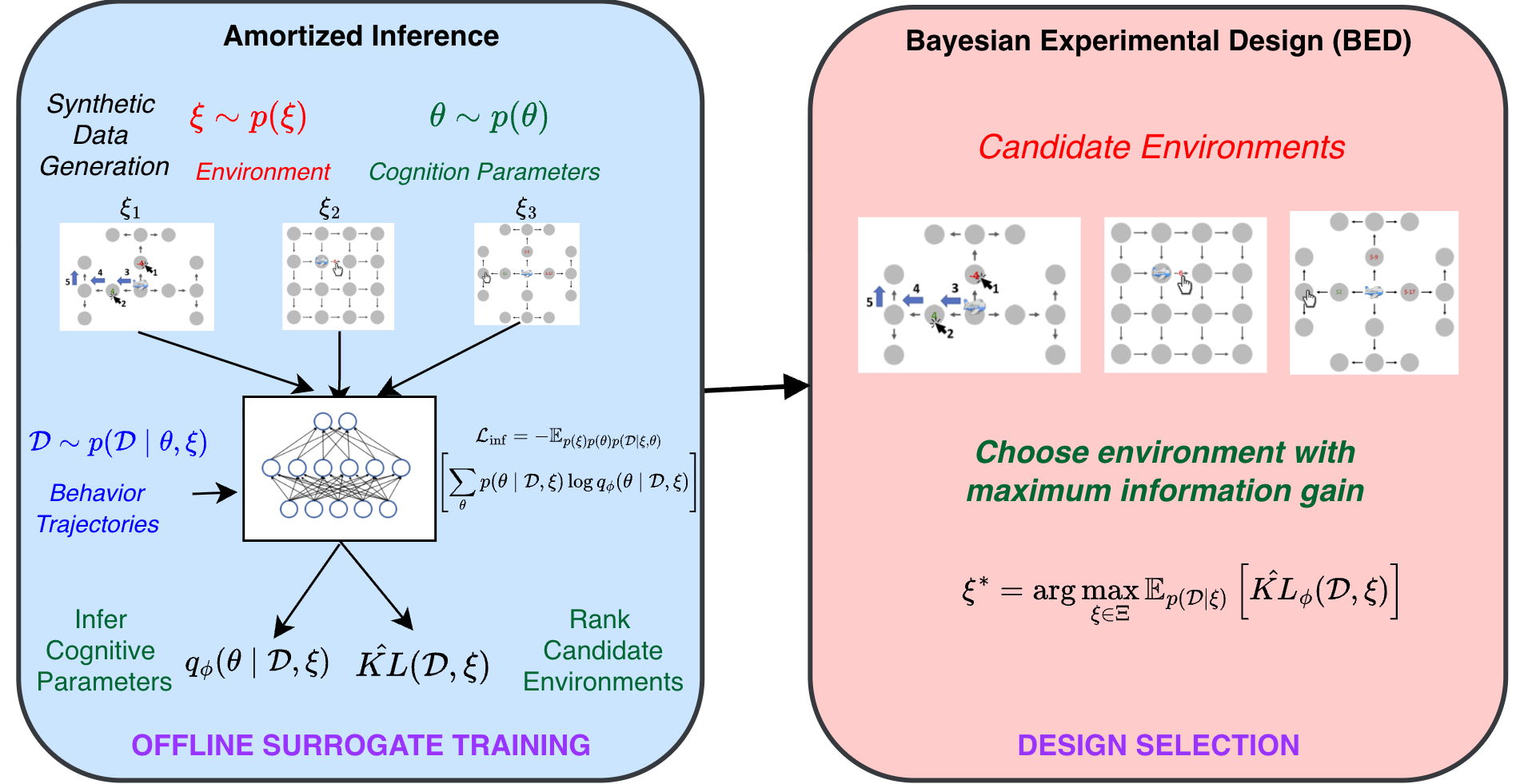}
    \caption{Overview of the proposed pipeline for identifying an informative environment for cognition parameter inference using Bayesian Experimental Design}
    \label{fig:overview}
\end{figure}

 Given a candidate design space $\Xi$, our goal is to identify experimental environments that maximize the information available for recovering latent cognitive parameters from observed behavior. To solve the environment-identification problem defined in Eq.~\ref{eq:best_eig}, we require a probabilistic model capable of simulating behavior under different latent cognitive parameters and evaluating the resulting information gained from candidate environments. To this end, we first define a generative behavioral simulator and the corresponding Bayesian posterior over latent cognitive parameters. We then formulate environment selection as a Bayesian Experimental Design (BED) problem by maximizing the expected information gain obtained from future behavioral observations. Finally, we introduce an amortized posterior approximation that enables efficient Bayesian Experimental Design without repeated posterior inference. Further, we use this amortized posterior approximation towards Amortized Bayesian experimental design for design evaluation, as depicted in Fig.~\ref{fig:overview}. 

\subsection{Behavioral Simulator}
\label{sec:behavioral_simulator}

To evaluate candidate environments, we require a generative model that simulates behaviour under different latent cognitive parameter settings. We model the agent using a parameterized planner with latent cognitive parameters $\theta \in \Theta$, which govern the agent's internal reasoning process. Given an environment $\xi$, the planner maintains true environment state $s_t$ and internal cognitive state $b_t$. As the agent interacts with the environment, the internal cognitive state is updated recursively according to latent cognitive parameters. Consequently, different parameter settings induce different planning behaviours while interacting with the same environment. The resulting internal cognitive state is then used to evaluate candidate actions, yielding action values \(Q_\theta(s_t,a;b_t)\). Actions are then sampled according to the softmax policy
\setlength{\abovedisplayskip}{0pt}
\setlength{\belowdisplayskip}{0pt}
\setlength{\abovedisplayshortskip}{0pt}
\setlength{\belowdisplayshortskip}{0pt}
\begin{equation}
\pi_\theta(a_t\mid s_t,b_t)
=
\frac{\exp\!\left(\eta Q_{\theta}(s_t,a_t;b_t)\right)}
{\sum_{a'}\exp\!\left(\eta Q_{\theta}(s_t,a';b_t)\right)},
\end{equation}
where $\eta$ is the inverse-temperature parameter controlling action stochasticity. This policy induces the trajectory likelihood
\setlength{\abovedisplayskip}{1pt}
\setlength{\belowdisplayskip}{1pt}
\setlength{\abovedisplayshortskip}{1pt}
\setlength{\belowdisplayshortskip}{1pt}
\begin{equation}
\label{eq:likelihood}
p(\mathcal D\mid\theta,\xi)
=
\prod_{i=1}^{N}\prod_{t=1}^{T_i}
\pi_\theta(a_t^{(i)}\mid s_t^{(i)},b_t^{(i)}),
\end{equation}
which defines the Bayesian posterior
\begin{equation}
p(\theta\mid\mathcal D,\xi)
\propto
p(\mathcal D\mid\theta,\xi)\,p(\theta).
\end{equation}

\subsection{Bayesian Experimental Design}
\label{sec:bed}
The behavioral simulator specifies the likelihood $p(\mathcal D|\theta,\xi)$ and therefore the Bayesian posterior over latent cognitive parameters. Substituting this posterior into the EIG objective from Eq.~\ref{eq:eig_entropy} yields the Bayesian Experimental Design criterion used to evaluate candidate environments that maximizes the expected information gained from future behavioral observations. Bayesian Experimental Design selects experimental environments by maximizing the expected information gained about the latent cognitive parameters from future observations. For a candidate design $\xi$, the information-theoretic objective introduced in Eq.~\ref{eq:eig_entropy} can be equivalently written as
\setlength{\abovedisplayskip}{1pt}
\setlength{\belowdisplayskip}{1pt}
\setlength{\abovedisplayshortskip}{1pt}
\setlength{\belowdisplayshortskip}{1pt}
\begin{equation}
\label{eq:eig_kl}
EIG(\xi)
=
\mathbb E_{p(\mathcal D \mid \xi)} 
\left[
KL\!\left(
p(\theta \mid \mathcal D,\xi)
\,\|\, 
p(\theta)
\right)
\right]
\end{equation}

\begin{equation}
\label{eq:eig_posterior}
EIG(\xi)
=
\mathbb E_{p(\theta)p(\mathcal D \mid \theta,\xi)}
\left[
\log \frac{p(\theta \mid \mathcal D,\xi)}{p(\theta)}
\right]
\end{equation}

These expressions define the expected information gain (EIG) objective. In this work, the expectation is estimated using Monte Carlo simulation, while the posterior $p(\theta\mid\mathcal D,\xi)$ is evaluated exactly for every simulated dataset and candidate
environment. We estimate the Bayesian Experimental Design objective using Monte Carlo samples from the generative model and exact posterior inference, which we refer as Monte Carlo Bayesian Experimental Design (MC-BED). Although posterior inference is exact, estimating EIG still requires repeated posterior evaluations over many simulated datasets, making design optimization computationally expensive.

\subsection{Amortized Posterior Inference}
\label{sec:amortized_posterior_inference}

Evaluating the Bayesian experimental design objective in Eq.~\ref{eq:eig_posterior}
requires repeated computation of the posterior
$p(\theta\mid\mathcal D,\xi)$ for many simulated datasets and candidate
environments. Since this posterior is obtained through Bayesian inference over
a simulator-defined likelihood, exact evaluation becomes the computational
bottleneck. To amortize this cost, we learn a neural posterior surrogate
$q_\phi(\theta\mid\mathcal D,\xi)$ that directly approximates the Bayesian
posterior from the observed trajectory dataset and the corresponding
environment. Training data are generated by sampling $\xi\sim p(\xi)$, $\theta\sim p(\theta)$, and $\mathcal D\sim p(\mathcal D\mid\theta,\xi)$. For each synthetic dataset, the corresponding Bayesian posterior is computed using the exact inference procedure and used as the supervision signal for the surrogate. The surrogate is supervised using exact Bayesian posterior distributions computed for synthetic datasets, effectively distilling the exact inference procedure into a neural posterior estimator.
\setlength{\abovedisplayskip}{0pt}
\setlength{\belowdisplayskip}{0pt}
\setlength{\abovedisplayshortskip}{0pt}
\setlength{\belowdisplayshortskip}{0pt}
\begin{equation}
\label{eq:training_obj}
\mathcal L_{\mathrm{inf}}
=
-
\mathbb E_{p(\xi)p(\theta)p(\mathcal D\mid\theta,\xi)}
\left[
\sum_{\theta}
p(\theta\mid\mathcal D,\xi)
\log q_\phi(\theta\mid\mathcal D,\xi)
\right]
\end{equation}

After training, posterior inference reduces to a single forward pass through
$q_\phi$, replacing repeated Bayesian inference with a reusable approximation
applicable across candidate environments. The surrogate receives the candidate
environment together with the observed trajectory dataset as input and outputs
an approximate posterior distribution over the latent cognitive parameters. \ref{eq:training_obj} defines the ideal amortized posterior inference objective. In our implementation, we augment this objective with additional supervision available during simulation. Since each synthetic training example provides both the exact Bayesian posterior and the ground-truth latent parameter, the posterior head is trained using a weighted combination of posterior distillation and supervised classification. Furthermore, to support efficient Bayesian experimental design, an auxiliary prediction head is trained to regress the exact posterior-to-prior KL used for environment ranking. Detailed architectural details are provided in the appendix.

\subsection{Amortized Bayesian Experimental Design}
\label{sec:abed}

Once the posterior surrogate has been learned, Amortized Bayesian Experimental Design (ABED) can be performed without repeatedly invoking exact Bayesian inference.
Specifically, we replace the exact posterior in
Eq.~\ref{eq:eig_posterior} with the amortized approximation
$q_\phi(\theta\mid\mathcal D,\xi)$,
\setlength{\abovedisplayskip}{1pt}
\setlength{\belowdisplayskip}{1pt}
\setlength{\abovedisplayshortskip}{1pt}
\setlength{\belowdisplayshortskip}{1pt}
\begin{equation}
\mathcal L_{\mathrm{ABED}}(\xi)
=
\mathbb E_{p(\theta)p(\mathcal D\mid\theta,\xi)}
\left[
\log
\frac{q_\phi(\theta\mid\mathcal D,\xi)}
{p(\theta)}
\right]
\end{equation}

The expectation within amortized objective is estimated using Monte Carlo samples,
\setlength{\abovedisplayskip}{0pt}
\setlength{\belowdisplayskip}{0pt}
\setlength{\abovedisplayshortskip}{0pt}
\setlength{\belowdisplayshortskip}{0pt}
\begin{equation}
\hat{\mathcal L}_{\mathrm{ABED}}(\xi)
\!\! = \!\!
\frac1N
\sum_{n=1}^{N}
\left[
\log
q_\phi(\theta_n\mid\mathcal D_n,\xi)
-
\log p(\theta_n)
\right]
\end{equation}

where $\theta_n\sim p(\theta)$, $\mathcal D_n\sim p(\mathcal D\mid\theta_n,\xi)$. 
Candidate environments are ranked according to $\hat{\mathcal L}_{\mathrm{ABED}}(\xi)$.
Unlike exact BED, the expensive posterior computation
is replaced by single neural-network evaluation, substantially reducing the
cost of evaluating candidate environments while preserving the Bayesian
information-theoretic objective.

\section{Related Work}

\paragraph{Bayesian Inverse Planning and Theory of Mind}
Bayesian inverse planning infers latent cognitive variables from observed action sequences using probabilistic generative models and Bayesian inference \cite{baker2007goal,baker2009action,griffiths2024bayesian}. Whereas most work focuses on inferring goals, beliefs, or preferences, we infer latent cognitive parameters governing resource-bounded planning \cite{griffiths2015rational,callaway2018resource}. Unlike inverse reinforcement learning, inverse planning explicitly models latent cognitive computations beyond reward functions, which are generally not recoverable through reward learning alone \cite{DBLP:conf/icml/ShahGAD19}. Recent work has extended inverse planning to richer cognitive models, including foundation-model-assisted inference over complex mental states \cite{gelpi2025towards,zhang2025autotom} and attention-aware inverse planning for estimating latent attentional biases \cite{banerjee2025estimating}. These advances improve the expressiveness of cognitive models, whereas our work addresses the complementary problem of selecting experimental environments that maximize information for latent cognitive parameter inference.

\paragraph{Bayesian Experimental Design and Active Environment Selection}
Bayesian experimental design (BED) provides a principled framework for selecting experiments that maximize expected information gain \cite{lindley1956measure,ryan2016review,foster2021deep,rainforth2024modern}. Since evaluating expected information gain requires repeated posterior inference, recent work has developed simulator-based, deep, and amortized BED methods that use approximate Bayesian inference and neural surrogates to enable scalable design optimization despite intractable likelihoods \cite{valentin2021bayesian,valentin2024designing,schreiber2024enhancing,musslick2025automating,valentin2024approximate,gautheron2026active,foster2021deep,huang2024amortized,ivanova2024automated,kennamer2023design}. Related work has also investigated amortized posterior inference for cognitive models \cite{radev2020amortized}, autonomous systems for closed-loop scientific discovery \cite{jagadish2026closing}, and experimental identifiability of Bayesian cognitive models \cite{hahn2025identifiability}. Complementing these directions, we formulate experimental environment selection itself as a Bayesian experimental design problem, combining amortized posterior inference with design-value prediction to efficiently identify informative environments for latent cognitive parameter inference in sequential decision-making.

\paragraph{Bounded Planning Models}
Planning depth is a central component of resource-bounded decision making and has been studied in domains including travelling salesman problems \cite{vickers2001human}, chess \cite{van2023expertise}, and grid-world navigation \cite{krusche2019adaptive,eluchans2025adaptive}. More recently, \cite{lancia2023humans} proposed an information-theoretic bounded-rational framework showing that human shortcut-finding balances travel-time savings against the cognitive cost of planning away from a learned default path.

\paragraph{Behavior-Diagnostic Game and Environment Design}
A closely related direction is behavior-diagnostic game design \cite{yang2019game}, which optimizes game environments to maximize mutual information between observed behavior and latent psychological traits using a variational objective. Complementary work extends the optimal game design paradigm of \cite{an2014optimally} to Bayesian experimental design for inverse planning in structured sequential decision environments. In contrast, we formulate environment selection itself as a Bayesian experimental design problem, maximizing posterior information gain for latent cognitive parameter inference.

\section{Experiments}
Our experiments address four research questions (RQ): 
\begin{itemize}
    \item[\textbf{RQ1:}] Which candidate environments permit reliable finite-sample recovery of latent cognitive parameters?
    \item[\textbf{RQ2:}] Does expected information gain identify environments that support accurate finite-sample recovery?
    \item[\textbf{RQ3:}] Can posterior inference and experimental-design evaluation be amortized while preserving useful environment rankings?
    \item[\textbf{RQ4:}] How does ABED compare to heuristic and learning-based methods?
    \item[\textbf{RQ5:}] Which structural and behavioral properties explain why some environments are more informative than others?
\end{itemize}
Together, these questions separate empirical recoverability, information-theoretic design quality, amortized approximation, and mechanistic explanation.

\begin{table*}[t]
\centering
\caption
{
Comparison of candidate Mouselab-MDP environments for planning-horizon inference with $H\in\{1,2,3\}$, fixed $K=2$, and five trajectories per dataset. RQ1 evaluates exact posterior recoverability and posterior separability, RQ2 evaluates Monte Carlo Bayesian Experimental Design (MC-BED), and RQ3 evaluates the amortized posterior surrogate and ABED. For readability, the main text reports the binary-tree benchmark; corresponding results for the expanded binary+ternary (\texttt{b2}+\texttt{b3}) design space are provided in the supplementary material. Bold indicates the best value in each column.
}
\label{tab:cross_method_results}
\setlength{\tabcolsep}{3.8pt}
\begin{tabular}{lcccccccccc}
\toprule
&
\multicolumn{4}{c}{\textbf{Exact Cognition Inference}}
&
\multicolumn{3}{c}{\textbf{Exact BED}}
&
\multicolumn{3}{c}{\textbf{Amortized BED}}
\\
\cmidrule(lr){2-5}
\cmidrule(lr){6-8}
\cmidrule(lr){9-11}

\textbf{Design}
&
\textbf{MAP}
&
\textbf{Ent.\ Red.}
&
\textbf{Min JS}
&
\textbf{Mean JS}
&
\textbf{Mean KL}
&
\textbf{$p(H_{\mathrm{true}})$}
&
\textbf{MAP}
&
\textbf{Surr.\ KL}
&
\textbf{$q(H_{\mathrm{true}})$}
&
\textbf{MAP}
\\
\midrule

\texttt{d2b2}
& 0.333
& $\approx 0$
& $\approx 0$
& $\approx 0$
& $\approx 0$
& 0.333
& 0.333
& 0.116
& 0.332
& 0.398
\\

\texttt{d3b2}
& 0.667
& 0.315
& 0.117
& 0.215
& 0.258
& 0.501
& 0.600
& 0.285
& 0.406
& 0.434
\\

\texttt{d4b2}
& 0.667
& 0.332
& 0.002
& 0.147
& 0.394
& 0.550
& 0.647
& 0.414
& \textbf{0.461}
& 0.542
\\

\texttt{d5b2}
& \textbf{1.000}
& 0.552
& \textbf{0.183}
& 0.361
& 0.495
& 0.623
& 0.693
& 0.448
& 0.445
& \textbf{0.551}
\\

\texttt{d6b2}
& 0.667
& \textbf{0.632}
& 0.001
& \textbf{0.442}
& \textbf{0.522}
& \textbf{0.636}
& \textbf{0.720}
& \textbf{0.533}
& 0.403
& 0.451
\\

\bottomrule
\end{tabular}
\end{table*}

\subsection{Experimental Setup}
\paragraph{Mouselab Environment}
We evaluate our framework on Mouselab-MDP~\cite{callaway2017mouselab}, a standard process-tracing paradigm for studying sequential planning and information acquisition in computational cognitive science. Agents sequentially reveal uncertain node rewards through clicks before terminating and selecting a root-to-leaf path, providing fine-grained behavioral evidence for Bayesian inverse planning. The tree-structured environment is both interpretable and computationally challenging, making it well suited for evaluating amortized Bayesian experimental design. Each candidate design is a tree $\xi=(d,b)$, where $d$ denotes the tree depth and $b$ the branching factor. We evaluate binary trees with $d\in\{2,\ldots,6\}$ and ternary trees with $d\in\{2,3,4\}$, denoted as \texttt{d\{depth\}b\{branch\}}. The root reward is fixed to zero, while each non-root node follows $R\sim\mathrm{Categorical}(\{-4,-2,2,4\})$, and every click incurs a cost of $-1$. Since the branching factor and reward distribution are fixed, the experiments isolate the effect of tree depth on planning-horizon recoverability. Exact recursive planning scales exponentially with planning horizon and rapidly becomes computationally expensive as the environment grows.

\paragraph{Cognitive Model}
Throughout this work, the latent cognitive parameters are planning horizon \(H\) and working-memory capacity \(K\). Planning horizon determines the depth of recursive lookahead, while working-memory capacity limits the amount of task-relevant information available during planning. The planner therefore evaluates future decisions using a bounded internal memory state of capacity \(K\), rather than the full history of observations. Working memory is implemented as a capacity-limited representation of the click history. When the number of revealed node values exceeds \(K\), the planner retains only the most recent \(K\) observations, while earlier observations are forgotten and treated as unobserved during subsequent planning. Consequently, smaller values of \(K\) induce greater information loss across sequential observations, whereas larger values support more persistent integration of previously acquired information.

\begin{table*}
\centering
\caption{
Comparison of candidate Mouselab-MDP environments for joint inference of
$\theta=(H,K)$ with $H\in\{1,2,3\}$ and $K\in\{2,5\}$.
RQ1 evaluates exact posterior recoverability, RQ2 evaluates Monte Carlo BED,
and RQ3 evaluates the amortized posterior surrogate and ABED on held-out Bank~C.
The surrogate is trained on Bank~A and selected on Bank~B.
Bold indicates the best value in each column.
}
\label{tab:results_h_k}
\setlength{\tabcolsep}{2.5pt}
\begin{tabular}{lcccccccccccc}
\toprule
&
\multicolumn{4}{c}{\textbf{RQ1: Exact Inference}}
&
\multicolumn{3}{c}{\textbf{RQ2: MC-BED}}
&
\multicolumn{5}{c}{\textbf{RQ3: Amortized Posterior \& ABED}}
\\
\cmidrule(lr){2-5}
\cmidrule(lr){6-8}
\cmidrule(lr){9-13}

Design
&
MAP$_{HK}$
&
Ent.\ Red.
&
Min JS
&
Mean JS
&
Mean KL
&
$p(\theta_{\rm true})$
&
MAP$_{HK}$
&
Pred.\ EIG
&
$q(\theta^*)$
&
MAP$_H$
&
MAP$_K$
&
MAP$_{HK}$
\\
\midrule

\texttt{d2b2}
& 0.167
& $\approx 0$
& $\approx 0$
& $\approx 0$
& $\approx 0$
& 0.167
& 0.167
& 0.034
& 0.167
& 0.308
& 0.533
& 0.192
\\

\texttt{d3b2}
& 0.667
& 0.942
& 0.007
& 0.484
& 0.974
& 0.534
& 0.650
& 1.008
& 0.380
& \textbf{0.592}
& 0.817
& \textbf{0.517}
\\

\texttt{d4b2}
& \textbf{1.000}
& \textbf{1.459}
& 0.070
& \textbf{0.625}
& 1.316
& 0.715
& 0.800
& 1.326
& 0.417
& 0.575
& \textbf{0.875}
& 0.492
\\

\texttt{d5b2}
& 0.833
& 1.306
& \textbf{0.083}
& 0.563
& \textbf{1.401}
& \textbf{0.784}
& \textbf{0.883}
& \textbf{1.446}
& \textbf{0.436}
& 0.533
& 0.867
& 0.458
\\

\bottomrule
\end{tabular}
\end{table*}

\paragraph{Training and Evaluation Protocol}

Trajectory data are generated once and partitioned into disjoint training, validation, and test splits. We consider five candidate environments and two inference settings: planning-horizon inference with \(H\in\{1,2,3\}\) (fixed \(K=2\)) and joint inference of \(\theta=(H,K)\). Exact posteriors and posterior-to-prior KL values are computed once for every training dataset and used as supervision for amortized learning. Model selection is performed on the validation split, and all reported results are averaged over five random seeds on held-out test datasets. The amortized surrogate predicts \(q_{\phi}(\theta\mid\mathcal{D},\xi)\) from three inputs: (i) environment descriptors, (ii) path-aware trajectory sequences encoded using a GRU with attention, and (iii) dataset-level behavioural summary statistics. The resulting representation is processed by two heads: a posterior head for latent cognitive inference and an auxiliary KL head for design evaluation. Training minimizes a multi-task objective combining posterior distillation, supervision from the ground-truth cognitive parameters, and regression of the exact posterior-to-prior KL, weighted by \(\alpha\) and \(\lambda\). We use learning rate of \(10^{-4}\), a batch size of 32, and 200 training epochs, with environments ranked by the predicted KL. More details in supplementary. 

\paragraph{Baselines}
\label{sec:baselines}

We compare ABED against three classes of alternatives: (i) \textit{structural heuristics} based solely on environment topology, (ii) \textit{behavioral heuristics} derived from simulated trajectories, and (iii) \textit{supervised regressors} that directly predict the Monte Carlo BED utility from environment descriptors and trajectory summaries. Unlike ABED, these methods produce only design scores and do not approximate the posterior over latent cognitive parameters. Posterior distillation and information-gain-aligned objectives are treated as ablations of the amortized inference model.

% \paragraph{Evaluation Metrics}

% We evaluate methods using three complementary criteria 1) \textit{Design ranking:}
% Spearman's $\rho$ and Kendall's $\tau$ with respect to the MC-BED ranking 2) \textit{Design selection:} Top-1 recovery and regret relative to the MC-BED optimum. 3) \textit{Posterior inference:} MAP accuracy, posterior probability assigned to the true planning horizon, and negative log-likelihood.

\subsection{RQ1: Can exact Bayesian inference reliably recover latent cognitive parameters?}

We first evaluate exact Bayesian inference on trajectory datasets generated under each candidate Mouselab-MDP environment. Posterior recovery is assessed using MAP accuracy, posterior entropy reduction, and pairwise Jensen--Shannon (JS) divergence, where the minimum pairwise JS divergence measures worst-case separability between latent cognitive parameter settings. Tables~\ref{tab:cross_method_results} and~\ref{tab:results_h_k} show that environment structure strongly affects posterior recoverability. The shallow design \texttt{d2b2} is consistently uninformative, whereas increasing environment depth generally improves recovery. However, no single environment is uniformly optimal across evaluation metrics. For planning-horizon inference, \texttt{d5b2} achieves highest MAP recovery, entropy reduction, and worst-case separability. For joint inference, \texttt{d4b2} yields the highest joint MAP recovery, whereas \texttt{d5b2} assigns the greatest posterior probability to true parameter pair. These results demonstrate that different notions of environment informativeness favour different designs, motivating Bayesian experimental design to select environments by maximizing expected information gain rather than any single posterior recovery metric.

\subsection{RQ2: Monte Carlo Bayesian Experimental Design}

Tables~\ref{tab:cross_method_results} and~\ref{tab:results_h_k} show that MC-BED evaluates candidate environments according to their expected information gain, providing a different criterion from the posterior recoverability metrics considered in RQ1. For planning-horizon inference (Table~\ref{tab:cross_method_results}), MC-BED identifies \texttt{d6b2} and \texttt{d5b2} as the most informative environments, even though \texttt{d5b2} achieves the highest exact MAP recovery. For joint inference (Table~\ref{tab:results_h_k}), \texttt{d5b2} maximizes both the expected information gain and the posterior probability assigned to the true parameter pair, whereas \texttt{d4b2} achieves the highest exact MAP recovery. These results demonstrate that maximizing expected information gain is not equivalent to maximizing finite-sample recovery. Rather than contradicting one another, the two objectives capture complementary notions of environment quality: exact inference evaluates how accurately latent cognitive parameters can be recovered after observing a finite dataset, whereas MC-BED selects environments expected to produce the greatest reduction in posterior uncertainty before any data are collected.

\subsection{RQ3: Can Amortized Bayesian Experimental Design efficiently recover informative environments?}

We evaluate whether the amortized surrogate reproduces the environment rankings obtained by Monte Carlo BED while replacing repeated Monte Carlo design evaluations with a single forward pass. Performance is assessed in terms of posterior recovery, predicted design utility, and agreement with the exact MC-BED ranking. Table~\ref{tab:cross_method_results} shows that the surrogate closely reproduces the MC-BED ranking, correctly identifying \texttt{d6b2} and \texttt{d5b2} as the most informative environments while consistently assigning the lowest utility to \texttt{d2b2}. Although posterior recovery is only moderate across environments, the predicted posterior-to-prior KL closely matches the exact MC-BED objective, indicating that accurate design evaluation does not require perfectly accurate posterior approximation. A similar pattern is observed for joint inference (Table~\ref{tab:results_h_k}), where ABED identifies the same optimal environment (\texttt{d5b2}) as MC-BED despite lower recovery of the joint cognitive parameter pair. Although posterior recovery is only moderate across environments, the predicted posterior-to-prior KL closely matches the exact MC-BED objective. The supplementary further reports exact cognition inference, MC-BED, and ABED results on an expanded design space comprising both binary (\texttt{b2}) and ternary (\texttt{b3}) Mouselab-MDP environments. Within the exact inference benchmark (five trajectories per dataset), \texttt{d5b2} and \texttt{d4b3} achieve the highest MAP recovery (1.00), while \texttt{d5b2} provides the largest worst-case separability (minimum JS = 0.183) and \texttt{d6b2} attains the greatest average entropy reduction (0.632). Within the MC-BED benchmark (ten trajectories per dataset), \texttt{d6b2} and \texttt{d5b2} are ranked as the two most informative environments, with mean KL values of 0.773 and 0.728, respectively. ABED maintains strong agreement with the exact MC-BED ranking (\(\rho=0.905\)) while correctly recovering the highest-ranked environment. These results indicate that the qualitative conclusions of the main paper remain unchanged under the expanded binary+ternary design space. On the evaluated Mouselab-MDP benchmark, preserving the relative information gain across candidate environments is sufficient to recover the correct design ranking, even when posterior inference remains imperfect. Consequently, ABED accurately approximates the MC-BED objective while replacing repeated exact posterior inference with a single forward pass.

\begin{figure}
    \centering \includegraphics[width=0.48\textwidth,keepaspectratio]{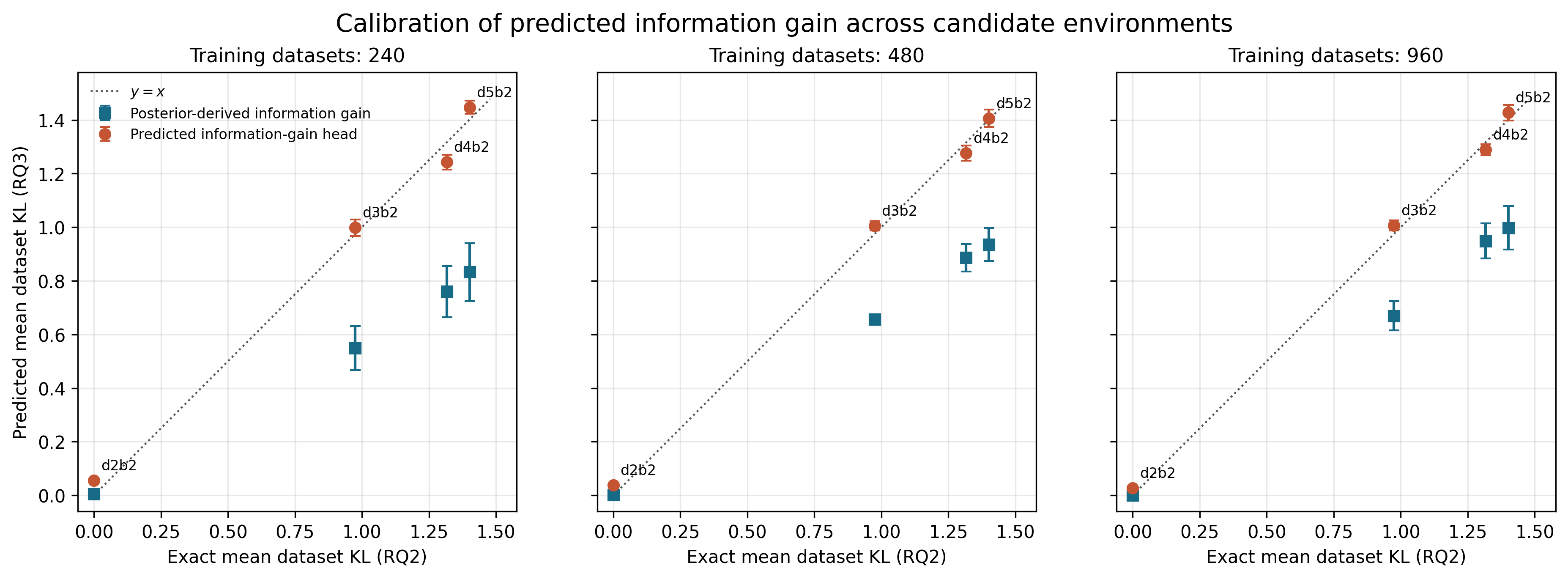}
    \caption{Calibration of predicted posterior-to-prior KL across candidate environments for joint inference of \((H,K)\) under different training-set sizes. The x-axis shows the exact MC-BED mean dataset KL and the y-axis the corresponding surrogate prediction. Blue squares denote KL computed from the amortized posterior, orange circles the auxiliary KL prediction head, and the dashed line indicates perfect calibration (\(y=x\))}
    \label{fig:ablation_calibration}
\end{figure}

\paragraph{Ablation} We evaluate three aspects of the amortized framework: the multi-task training objective, the trajectory representation, and the amount of training data. Across all settings, the auxiliary KL prediction head consistently recovers the MC-BED ranking, whereas posterior recovery is more sensitive to the choice of supervision and representation. For five trajectories per parameter-design pair, varying the objective weights has little effect on posterior inference, with \(H\)-MAP accuracy remaining stable across all settings. In contrast, explicit KL supervision consistently recovers the exact BED ranking with zero regret, whereas design selection based on posterior-derived KL is less reliable. Increasing the training set beyond 480 datasets provides only modest improvements, indicating that design-value prediction saturates with moderate training data. Figure~\ref{fig:ablation_calibration} further shows that the auxiliary KL head remains well calibrated across training-set sizes, closely matching the exact posterior-to-prior KL, while KL computed from the amortized posterior systematically underestimates the information gain, particularly for the most informative environments. Detailed ablation is in appendix.

\begin{table*}
\centering
\caption{Comparison of ABED with heuristic and learning-based methods for identifying informative environments. Methods are compared in terms of ranking performance and support for posterior inference, MC-BED objective prediction, or both. Spearman's \(\rho\) measures agreement with exact MC-BED ranking, and regret is computed relative to exact MC-BED optimum}
\label{tab:baseline_comparison}
\setlength{\tabcolsep}{4pt}
\begin{tabular}{lcccccc}
\toprule
\textbf{Method} &
\textbf{Posterior} &
\textbf{MC-BED} &
\textbf{Both Tasks} &
\textbf{Spearman's $\rho$} &
\textbf{Best} &
\textbf{Regret} \\
\midrule

Largest tree
& $\times$
& Heuristic
& $\times$
& 1.000
& \texttt{d5b2}
& 0.000 \\

Behavioural diversity
& $\times$
& Heuristic
& $\times$
& 0.800
& \texttt{d4b2}
& 0.085 \\

Linear regression
& $\times$
& $\checkmark$
& $\times$
& 1.000
& \texttt{d5b2}
& 0.000 \\

Logistic posterior
& $\checkmark$
& $\times$
& $\times$
& 0.800
& \texttt{d4b2}
& 0.085 \\

\midrule

\textbf{ABED}
& \textbf{$\checkmark$}
& \textbf{$\checkmark$}
& \textbf{$\checkmark$}
& \textbf{1.000}
& \textbf{\texttt{d5b2}}
& \textbf{0.000} \\

\bottomrule
\end{tabular}
\end{table*}

\begin{table}
\centering
\caption{
Best-performing environments under three experimental objectives for each
latent cognitive state $\theta=(H,K)$. Entries report the optimal design
followed by the corresponding metric value.
}
\label{tab:rq5}
\setlength{\tabcolsep}{3pt}
\begin{tabular}{lccc}
\toprule
\textbf{True $\theta$}
&
\shortstack{\textbf{Information}\\(KL)}
&
\shortstack{\textbf{Efficiency}\\(KL/click)}
&
\shortstack{\textbf{Recovery}\\(MAP)}
\\
\midrule

$(1,2)$ &
\texttt{d5b2} {\scriptsize (1.574)} &
\texttt{d3b2} {\scriptsize (0.0454)} &
\texttt{d2b2} {\scriptsize (1.000)}
\\

$(1,5)$ &
\texttt{d5b2} {\scriptsize (1.512)} &
\texttt{d4b2} {\scriptsize (0.0276)} &
\texttt{d5b2} {\scriptsize (0.950)}
\\

$(2,2)$ &
\texttt{d5b2} {\scriptsize (1.223)} &
\texttt{d3b2} {\scriptsize (0.0271)} &
\texttt{d5b2} {\scriptsize (0.900)}
\\

$(2,5)$ &
\texttt{d5b2} {\scriptsize (1.405)} &
\texttt{d3b2} {\scriptsize (0.0242)} &
\texttt{d5b2} {\scriptsize (0.900)}
\\

$(3,2)$ &
\texttt{d5b2} {\scriptsize (1.342)} &
\texttt{d3b2} {\scriptsize (0.0308)} &
\texttt{d5b2} {\scriptsize (0.900)}
\\

$(3,5)$ &
\texttt{d4b2} {\scriptsize (1.368)} &
\texttt{d4b2} {\scriptsize (0.0236)} &
\texttt{d4b2}/\texttt{d5b2} {\scriptsize (0.750)}
\\

\bottomrule
\end{tabular}
\end{table}

\subsection{RQ4: How does ABED compare to heuristic and learning-based methods?}

Table~\ref{tab:baseline_comparison} compares ABED with structural heuristics, behavioural heuristics, supervised regressors, and posterior-based baselines. These methods differ not only in ranking performance but also in the inference capabilities they provide. Structural heuristics rely solely on environment properties, supervised regressors directly predict the MC-BED objective without performing posterior inference, and logistic-posterior inference estimates cognitive parameters without explicitly predicting design utility. In contrast, ABED jointly learns an amortized posterior \(q_\phi(\theta\mid D,\xi)\) together with a design-value predictor, enabling both cognitive parameter inference and environment evaluation within a single model.

For our setup, MC-BED informativeness increases approximately monotonically with environment depth, allowing the largest-tree heuristic and supervised regressors to recover the optimal environment. Behavioural diversity and logistic-posterior inference are less reliable, selecting the suboptimal design \texttt{d4b2} and incurring an MC-BED regret of \(0.085\). ABED also recovers the exact MC-BED ranking with zero regret while simultaneously supporting posterior inference, making it the only approach that performs both tasks within a unified amortized framework. The planning-horizon benchmark reported in the Appendix exhibits the same qualitative behaviour.

\subsection{RQ5: Why do different environments become informative?}

Table~\ref{tab:rq5} summarizes the environments that maximize conditional expected information gain, information efficiency (KL/click), and parameter recovery for each latent cognitive state. The results reveal that environment informativeness depends jointly on the underlying cognitive state and the experimental objective. Deeper environments provide more opportunities to distinguish complex planning strategies and therefore maximize conditional expected information gain for most latent states, with \texttt{d5b2} being optimal in five of the six cases. In contrast, shallower environments such as \texttt{d3b2} (and occasionally \texttt{d4b2}) achieve higher information efficiency because they require fewer participant interactions to obtain each unit of information. The environment that maximizes finite-sample recovery is also not always the one that maximizes expected information gain. For example, when $(H,K)=(1,2)$, the shallow environment \texttt{d2b2} achieves perfect MAP recovery despite providing substantially less expected information than \texttt{d5b2}. This occurs because simple cognitive strategies can often be identified reliably in small environments, whereas deeper environments are more valuable for distinguishing more complex behaviours. Overall, these results demonstrate that no single environment is universally optimal: the preferred design depends on the latent cognitive state and on whether the goal is to maximize expected information, recovery accuracy, or experimental efficiency.

\subsection{Implications for Adaptive Cognitive Experimental Design}
The experimental results collectively show that no single environment is universally optimal for cognitive parameter inference. RQ1 and RQ2 demonstrate that posterior recoverability and expected information gain favour different environments, while RQ5 shows that the optimal design further depends on the latent cognitive state. Together, these findings motivate adaptive environment selection conditioned on the current posterior over cognitive parameters rather than relying on a fixed experimental environment. ABED provides a natural foundation for adaptive Bayesian experimental design by jointly estimating the posterior over latent cognitive parameters and predicting the information gain of candidate environments. This enables efficient environment evaluation through a single forward pass instead of repeated Monte Carlo BED computations. Although adaptive environment selection is beyond the scope of this work, the proposed framework provides the components required for sequential experiment selection conditioned on an evolving posterior over latent cognitive parameters.

\section{Conclusion}

We formulated environment selection for Bayesian cognition parameter inference as a Bayesian Experimental Design (BED) problem, treating sequential decision environments as experimental designs rather than fixed evaluation settings. We established an exact Monte Carlo BED benchmark and introduced an amortized BED (ABED) framework that jointly performs posterior inference and design evaluation. Experiments on Mouselab-MDP demonstrate that environment structure strongly influences cognitive parameter inference, and that no single environment is uniformly optimal across posterior recoverability, expected information gain, and participant effort. ABED accurately recovers the MC-BED environment ranking while substantially reducing the cost of repeated posterior estimation. Although simple depth-based heuristics perform well on the restricted benchmark considered here, this reflects the limited design space rather than a property of the proposed framework. In contrast, ABED does not rely on structural regularities and naturally extends to settings where environment informativeness cannot be inferred from topology alone. These findings suggest that posterior-guided environment selection is a promising direction for scalable and participant-efficient cognitive assessment, providing a principled foundation for designing informative sequential decision-making environments for Bayesian cognition parameter inference. More broadly, our work argues that experimental environments should themselves be optimized as part of cognitive model inference.

\section{Acknowledgement}
This work was supported by a UKRI Turing AI World Leading Researcher Fellowship on AI for Person-Centred and Teachable Autonomy (grant EP/Z534833/1)

\bibliography{aaai2027}


% \clearpage
% \appendix

\clearpage
\appendix

\begin{center}
{\Huge\bfseries Supplementary Material}

\vspace{0.2cm}

\end{center}

\section{Experimental Setup}

\subsection{Cognitive Model}

\paragraph{Benchmark Design Space}

The experiments consider a controlled family of symmetric Mouselab-MDP trees that vary in depth and branching factor. This restricted design space permits exact Bayesian inference and exhaustive comparison against MC-BED, providing a reliable benchmark for evaluating amortized design selection. Within the binary benchmark, expected information gain increases approximately with tree depth, making simple topology-based heuristics unexpectedly competitive. The expanded binary--ternary benchmark partially mitigates this effect by introducing environments with similar depth but different branching factors, leading to different posterior recoverability and MC-BED rankings.

\paragraph{Environment and behaviour}
We use symmetric Mouselab-MDP trees parameterized by depth $d$ and branching factor $b$. The root reward is fixed at zero, non-root rewards follow $R\sim\mathrm{Categorical}(\{-4,-2,2,4\})$, and each click incurs a cost of $-1$. A trajectory ends when the agent terminates or reaches the 50-action limit. The latent cognitive parameters are planning horizon $H$ and working-memory capacity $K$. Planning horizon determines recursive lookahead, with $V(s,h,K)=\max_a Q(s,a,h,K)$, termination value $Q(s,\mathrm{terminate},h,K)=\mathbb{E}[R_{\mathrm{terminal}}\mid s_K]$, and click value $Q(s,\mathrm{click}\;j,h,K)=-1+\mathbb{E}_{s'}[V(s',h-1,K)]$. Working memory retains at most $K$ revealed node values; observations beyond capacity revert to their prior distributions during later planning. Planning-horizon experiments fix $K=2$, while joint inference uses $K\in\{2,5\}$. Actions are sampled from the softmax policy with $\eta=2.0$. Only actions feasible in the true environment are available, so previously revealed nodes cannot be clicked again even if forgotten by the cognitive model.

\subsection{Trajectory Generation and Data Splits}

For each environment $\xi$ and latent configuration $\theta=(H,K)$, trajectories are generated using deterministic random seeds. At each step, the simulator reconstructs the memory-limited belief state, evaluates termination and feasible click actions, samples an action, records the pre-action state and action, and updates the environment. Trajectories are generated once and partitioned into disjoint training, validation, and test sets. Dataset examples group a fixed number of trajectories from the same $(\xi,H,K)$ configuration, ensuring that evaluation trajectories are not used for surrogate training. We consider three settings: planning-horizon inference on binary trees with $H\in\{1,2,3\}$ and fixed $K=2$; an expanded planning-horizon benchmark containing binary and ternary trees; and joint inference with $H\in\{1,2,3\}$ and $K\in\{2,5\}$.

\subsection{Exact Bayesian Inference}

Exact posterior inference is used for both recoverability analysis and Monte Carlo BED. For each candidate $\theta$, the observed trajectories are replayed under the corresponding cognitive model, giving $\log p(\mathcal D\mid\theta,\xi)=\sum_{\tau\in\mathcal D}\sum_t\log\pi_\theta(a_t^\tau\mid s_t^\tau,\xi)$. The posterior is computed as $p(\theta\mid\mathcal D,\xi)\propto p(\mathcal D\mid\theta,\xi)p(\theta)$ using log-space normalization. We use a uniform prior over $H\in\{1,2,3\}$ for planning-horizon inference and over the six $(H,K)$ combinations for joint inference. Recoverability is measured using MAP accuracy, entropy reduction, minimum pairwise Jensen--Shannon divergence, and mean pairwise Jensen--Shannon divergence. Minimum JS measures worst-case separability, while mean JS summarizes average posterior separation.

\begin{table*}[t]
\centering

\begin{minipage}[t]{0.49\textwidth}
\centering
\caption{Exact recoverability for planning-horizon inference with
$H\in\{1,2,3\}$ and fixed $K=2$. Designs are ordered by minimum pairwise JS.}
\label{tab:stage1_honly}
\small
\setlength{\tabcolsep}{2.2pt}
\begin{tabular}{lccccc}
\toprule
\textbf{Design} &
\textbf{MAP} &
\textbf{Ent.\ Red.} &
\textbf{Min JS} &
\textbf{Mean JS} &
\textbf{Hardest Pair} \\
\midrule
\texttt{d5b2} & \textbf{1.000} & 0.753 & \textbf{0.394} & 0.507 &
$H=1$ vs.\ $H=2$ \\
\texttt{d6b2} & \textbf{1.000} & \textbf{0.861} & 0.338 &
\textbf{0.572} & $H=2$ vs.\ $H=3$ \\
\texttt{d4b2} & 0.667 & 0.596 & 0.105 & 0.383 &
$H=2$ vs.\ $H=3$ \\
\texttt{d3b2} & 0.667 & 0.405 & 0.076 & 0.277 &
$H=2$ vs.\ $H=3$ \\
\texttt{d2b2} & 0.333 & $\approx 0$ & $\approx 0$ & $\approx 0$ &
$H=1$ vs.\ $H=2$ \\
\bottomrule
\end{tabular}
\end{minipage}
\hfill
\begin{minipage}[t]{0.49\textwidth}
\centering
\caption{Exact recoverability for joint inference of $(H,K)$ with
$H\in\{1,2,3\}$ and $K\in\{2,5\}$. Designs are ordered by minimum pairwise JS.}
\label{tab:stage1_joint}
\small
\setlength{\tabcolsep}{2.2pt}
\begin{tabular}{lccccc}
\toprule
\textbf{Design} &
\textbf{MAP} &
\textbf{Ent.\ Red.} &
\textbf{Min JS} &
\textbf{Mean JS} &
\textbf{Hardest Pair} \\
\midrule
\texttt{d5b2} & 0.833 & 1.306 & \textbf{0.083} & 0.563 &
{\tiny $(H{=}2,K{=}5)$ vs.\ $(H{=}3,K{=}5)$} \\
\texttt{d4b2} & \textbf{1.000} & \textbf{1.459} & 0.070 &
\textbf{0.625} &
{\tiny $(H{=}2,K{=}2)$ vs.\ $(H{=}3,K{=}2)$} \\
\texttt{d3b2} & 0.667 & 0.942 & 0.007 & 0.484 &
{\tiny $(H{=}1,K{=}5)$ vs.\ $(H{=}2,K{=}5)$} \\
\texttt{d2b2} & 0.167 & $\approx 0$ & $\approx 0$ & $\approx 0$ &
{\tiny $(H{=}1,K{=}2)$ vs.\ $(H{=}1,K{=}5)$} \\
\bottomrule
\end{tabular}
\end{minipage}

\end{table*}

\subsection{Amortized Posterior and Design-Value Estimation}

\paragraph{Representation and architecture}
Each example contains an environment $\xi$, trajectory dataset $\mathcal D$, generating parameter $\theta$, exact posterior target, and exact posterior-to-prior KL target. The model combines environment descriptors, trajectory-level sequence features, and dataset-level behavioural summaries. We compare three representations. \textit{Basic} uses structural action tokens and aggregate summaries. \textit{Path-aware} adds ancestor, sibling, root-branch, revealed-node, and best-path evidence. \textit{Raw-state} additionally encodes padded belief states recurrently. Environment descriptors and summary features are encoded by separate two-layer multilayer perceptrons. Each trajectory is encoded by a GRU with timestep attention, and trajectory embeddings are aggregated using attention and mean pooling. The fused representation feeds a categorical posterior head $q_\phi(\theta\mid\mathcal D,\xi)$ and a non-negative scalar head predicting posterior-to-prior KL. For joint inference, separate logits over $H$ and $K$ are combined into a normalized joint posterior.

\paragraph{Training objective}
Let $p_i(\theta)$ denote the exact posterior target, $y_i$ the generating latent state, and $q_{\phi,i}(\theta)$ the predicted posterior. The posterior loss is $\mathcal L_{\mathrm{post}}=\alpha[-\sum_\theta p_i(\theta)\log q_{\phi,i}(\theta)]+(1-\alpha)[-\log q_{\phi,i}(y_i)]$, combining posterior distillation and hard-label supervision. Given exact KL target $\kappa_i$ and scalar prediction $k_{\phi,i}$, the full objective is $\mathcal L
=
\mathcal L_{\mathrm{post}}
+
\lambda\,\mathcal L_{\mathrm{reg}}(k_{\phi,i},\kappa_i)$. When $\lambda>0$, environments are ranked by the mean scalar KL prediction; when $\lambda=0$, ranking uses $\mathrm{KL}(q_\phi(\theta\mid\mathcal D,\xi)\|p(\theta))$. The former directly learns the design-evaluation target, while the latter depends on posterior approximation quality.

\paragraph{Hyperparamter Settings}
Here, we use a hidden dimension of 128, a dropout rate of 0.1, batch size of 32, and is trained for 150--600 epochs during model development and 200 epochs in the systematic ablation studies. The learning rate is set to $10^{-4}$ for planning-horizon inference and $3\times10^{-4}$ for joint $(H,K)$ inference. We evaluate posterior-distillation weights $\alpha\in\{0,0.25,0.5,0.75,1\}$ and KL-loss weights $\lambda\in\{0,0.1,0.2,0.5,1\}$ across four feature representations: \textit{basic}, \textit{path-aware} and \textit{raw-state}. Unless otherwise stated, results are averaged over five random model seeds $\{7,17,49,2026,2027\}$, while a fixed data-generation seed of 2025 is used throughout the systematic ablations. Experiments were conducted on Linux compute servers equipped with Intel Xeon processors and NVIDIA GeForce GTX 1080/1080 Ti GPUs (8--11\,GB GPU memory) using CUDA~12.2.

\section{Additional Results}

\subsection{Exact Posterior Recoverability}

Tables~\ref{tab:stage1_honly} and~\ref{tab:stage1_joint} report MAP recovery,
entropy reduction, and pairwise Jensen--Shannon (JS) divergence. Minimum JS
measures worst-case separability, whereas mean JS summarizes average
separation across latent states. Joint inference expands the hypothesis space from three horizons to six
$(H,K)$ combinations, reducing worst-case separability and overall recovery. In both settings, deeper environments are substantially more informative than the shallow \texttt{d2b2} control.

\subsection{Expanded Binary-Ternary Benchmark}
\label{sec:expanded_b2b3}

We extend the design space to include both binary and ternary Mouselab-MDP trees to evaluate whether the conclusions from the binary benchmark generalize to more diverse environment structures. Table~\ref{tab:cross_method_results_b2b3} summarizes all three RQs of the proposed pipeline. RQ1 evaluates exact posterior recoverability, RQ2 reports exact MC-BED utilities (Table \ref{tab:stage2_b2b3}), and RQ3 evaluates the amortized surrogate. Consistent with the binary benchmark, shallow environments provide little posterior separation, whereas deeper environments substantially improve inference quality. However, RQ1 and RQ2 emphasize different notions of informativeness. Designs such as \texttt{d5b2} maximize worst-case posterior separability, while \texttt{d6b2} maximizes expected information gain despite exhibiting a much smaller minimum pairwise JS divergence. This demonstrates that posterior recoverability and Bayesian experimental design optimize related but distinct objectives. The amortized surrogate closely reproduces the RQ2 ordering, achieving Spearman $\rho=0.952$, perfect top-one and top-two recovery, and zero design-selection regret. Although posterior recovery remains imperfect ($H$-MAP $=0.537$), the learned KL head accurately predicts environment value, indicating that reliable design ranking is substantially easier than recovering the full posterior.

% \begin{table*}[t]
% \centering
% \caption{Exact posterior recoverability for planning-horizon inference ($H\in\{1,2,3\}$, fixed $K=2$) for binary and ternary trees. Designs are ordered by depth}
% \label{tab:stage1_honly_b2b3}
% \setlength{\tabcolsep}{4pt}
% \begin{tabular}{lcccccc}
% \toprule
% \textbf{Design} &
% \textbf{MAP} &
% \textbf{Ent.\ Red.} &
% \textbf{Min JS} &
% \textbf{Mean JS} &
% \textbf{Hardest Pair} \\
% \midrule
% \texttt{d2b2} & 0.333 & 0.000 & 0.000 & 0.000 & $H{=}1,K{=}2$ vs.\ $H{=}2,K{=}2$ \\
% \texttt{d2b3} & 0.333 & 0.034 & 0.005 & 0.020 & $H{=}1,K{=}2$ vs.\ $H{=}2,K{=}2$ \\
% \texttt{d3b2} & 0.667 & 0.510 & 0.145 & 0.358 & $H{=}2,K{=}2$ vs.\ $H{=}3,K{=}2$ \\
% \texttt{d3b3} & 0.333 & 0.327 & 0.065 & 0.239 & $H{=}1,K{=}2$ vs.\ $H{=}3,K{=}2$ \\
% \texttt{d4b2} & 1.000 & 0.542 & 0.168 & 0.368 & $H{=}2,K{=}2$ vs.\ $H{=}3,K{=}2$ \\
% \texttt{d4b3} & 1.000 & 0.835 & 0.387 & 0.547 & $H{=}2,K{=}2$ vs.\ $H{=}3,K{=}2$ \\
% \texttt{d5b2} & 1.000 & 0.893 & \textbf{0.420} & \textbf{0.583} & $H{=}2,K{=}2$ vs.\ $H{=}3,K{=}2$ \\
% \texttt{d6b2} & 0.667 & \textbf{0.935} & 0.007 & 0.463 & $H{=}2,K{=}2$ vs.\ $H{=}3,K{=}2$ \\
% \bottomrule
% \end{tabular}
% \end{table*}

\begin{table}[t]
\centering
\caption{
Exact MC-BED results for planning-horizon inference with
$H\in\{1,2,3\}$, fixed $K=2$, across binary and ternary trees,
using ten trajectories per dataset.
}
\label{tab:stage2_b2b3}
\setlength{\tabcolsep}{4pt}
\begin{tabular}{lcccc}
\toprule
\textbf{Design} &
\textbf{Mean KL} &
\textbf{SD} &
\textbf{$p(H_{\rm true})$} &
\textbf{MAP} \\
\midrule
\texttt{d6b2} & \textbf{0.739} & 0.286 & 0.763 & 0.827 \\
\texttt{d5b2} & 0.670 & 0.300 & 0.727 & 0.787 \\
\texttt{d4b3} & 0.663 & 0.282 & \textbf{0.768} & \textbf{0.853} \\
\texttt{d4b2} & 0.566 & 0.270 & 0.665 & 0.720 \\
\texttt{d3b2} & 0.409 & 0.243 & 0.600 & 0.667 \\
\texttt{d3b3} & 0.396 & 0.241 & 0.564 & 0.680 \\
\texttt{d2b3} & 0.052 & 0.066 & 0.360 & 0.453 \\
\texttt{d2b2} & 0.000 & 0.000 & 0.333 & 0.333 \\
\bottomrule
\end{tabular}
\end{table}

\begin{table*}[t]
\centering
\caption{
Comparison of eight candidate Mouselab-MDP environments for
planning-horizon inference with $H\in\{1,2,3\}$, fixed $K=2$, and
ten trajectories per dataset. RQ1 evaluates exact posterior
recoverability and posterior separability. RQ2 evaluates exact
Monte Carlo Bayesian experimental design on the held-out Bank-C
benchmark, using 25 Monte Carlo datasets per true parameter value.
RQ3 evaluates the amortized posterior surrogate trained on Bank-A
targets estimated using 50 Monte Carlo datasets per true parameter
value. Bold denotes the
best value in each column
}
\label{tab:cross_method_results_b2b3}
\setlength{\tabcolsep}{4pt}
\begin{tabular}{lcccccccccc}
\toprule
&
\multicolumn{4}{c}{\textbf{Exact Cognition Inference (RQ1)}}
&
\multicolumn{3}{c}{\textbf{Exact BED (RQ2)}}
&
\multicolumn{3}{c}{\textbf{Amortized BED (RQ3)}}
\\
\cmidrule(lr){2-5}
\cmidrule(lr){6-8}
\cmidrule(lr){9-11}

\textbf{Design}
&
\textbf{MAP}
&
\textbf{Ent.\ Red.}
&
\textbf{Min JS}
&
\textbf{Mean JS}
&
\textbf{Mean KL}
&
\textbf{$p(H_{\mathrm{true}})$}
&
\textbf{MAP}
&
\textbf{Surr.\ KL}
&
\textbf{$q(H_{\mathrm{true}})$}
&
\textbf{MAP}
\\
\midrule

\texttt{d2b2}
& 0.333
& 0.000
& 0.000
& 0.000
& 0.000
& 0.333
& 0.333
& 0.106
& 0.335
& 0.307
\\

\texttt{d3b2}
& 0.667
& 0.510
& 0.145
& 0.358
& 0.409
& 0.600
& 0.667
& 0.394
& 0.491
& 0.600
\\

\texttt{d4b2}
& \textbf{1.000}
& 0.542
& 0.168
& 0.368
& 0.566
& 0.665
& 0.720
& 0.596
& \textbf{0.549}
& 0.600
\\

\texttt{d5b2}
& \textbf{1.000}
& 0.893
& \textbf{0.420}
& \textbf{0.583}
& 0.670
& 0.727
& 0.787
& 0.645
& 0.527
& 0.600
\\

\texttt{d6b2}
& 0.667
& \textbf{0.935}
& 0.007
& 0.463
& \textbf{0.739}
& 0.763
& 0.827
& \textbf{0.688}
& 0.490
& 0.493
\\

\midrule

\texttt{d2b3}
& 0.333
& 0.034
& 0.005
& 0.020
& 0.052
& 0.360
& 0.453
& 0.103
& 0.346
& 0.440
\\

\texttt{d3b3}
& 0.333
& 0.327
& 0.065
& 0.239
& 0.396
& 0.564
& 0.680
& 0.495
& 0.479
& 0.587
\\

\texttt{d4b3}
& \textbf{1.000}
& 0.835
& 0.387
& 0.547
& 0.663
& \textbf{0.768}
& \textbf{0.853}
& 0.638
& 0.544
& \textbf{0.667}
\\

\bottomrule
\end{tabular}
\end{table*}

% \begin{table}[t]
% \centering
% \caption{Preliminary comparison between exact Stage 2 and Stage 3 design rankings}
% \label{tab:stage3_rank}
% \begin{tabular}{lcc}
% \toprule
% Design & Stage 2 & Stage 3\\
% \midrule
% \texttt{d6b2} & 1 & 1\\
% \texttt{d5b2} & 2 & 3\\
% \texttt{d4b3} & 3 & 2\\
% \texttt{d4b2} & 4 & 4\\
% \texttt{d3b2} & 5 & 6\\
% \texttt{d3b3} & 6 & 5\\
% \texttt{d2b3} & 7 & 8\\
% \texttt{d2b2} & 8 & 7\\
% \bottomrule
% \end{tabular}
% \end{table}

\subsection{Per-Parameter Joint Inference}

Table~\ref{tab:joint_theta_summary} reports exact performance pooled across the four binary environments and grouped by the generating latent state. Recovery is strongest for $(H,K)=(1,2)$. Increasing $K$ reduces recovery for
$H=1$, while MAP remains near $0.53$--$0.55$ for $H\in\{2,3\}$. Residual
confusion is therefore concentrated mainly between adjacent planning horizons,
rather than reflecting a uniform failure to distinguish memory capacities.

\begin{table}[t]
\centering
\caption{Exact joint-inference performance pooled across four binary
environments}
\label{tab:joint_theta_summary}
\setlength{\tabcolsep}{3.5pt}
\begin{tabular}{cccccc}
\toprule
\textbf{$H$} &
\textbf{$K$} &
\textbf{Mean KL} &
\textbf{$p(\theta_{\rm true})$} &
\textbf{Log Ratio} &
\textbf{MAP} \\
\midrule
1 & 2 & \textbf{1.072} & \textbf{0.687} & \textbf{1.171} &
\textbf{0.950} \\
1 & 5 & 0.957 & 0.585 & 0.954 & 0.637 \\
2 & 2 & 0.851 & 0.481 & 0.873 & 0.550 \\
2 & 5 & 0.848 & 0.511 & 0.898 & 0.550 \\
3 & 2 & 0.894 & 0.515 & 0.870 & 0.537 \\
3 & 5 & 0.917 & 0.521 & 0.885 & 0.525 \\
\bottomrule
\end{tabular}
\end{table}

\subsection{Statistical Uncertainty and Seed Variability}
\label{sec:uncertainty}

We distinguish two sources of uncertainty. RQ2 uncertainty reflects Monte Carlo variability in estimating expected information gain, whereas RQ3 uncertainty reflects variability across independently trained amortized surrogates. Since all RQ3 models use the same fixed RQ2 teacher targets, RQ2 uncertainty is estimated from held-out Monte Carlo datasets, while RQ3 uncertainty is estimated across five training seeds (\{7,17,49,2026,2027\}). Because only five seeds are used, the reported confidence intervals should be interpreted as uncertainty estimates rather than formal significance tests.

\paragraph{RQ2 Monte Carlo uncertainty:} For the joint $(H,K)$ benchmark ($H\in\{1,2,3\}$, $K\in\{2,5\}$), each design is evaluated using 120 held-out Monte Carlo datasets (20 per latent parameter setting). Table~\ref{tab:stage2_uncertainty} reports the mean expected information gain, dataset-level standard deviation, and 95\% confidence interval. The leading designs remain distinguishable: the mean KL difference between \texttt{d5b2} and \texttt{d4b2} is $0.085$ with an approximate 95\% confidence interval of $[0.009,0.161]$, indicating that the advantage of \texttt{d5b2} exceeds Monte Carlo variability.

\begin{table}[t]
\centering
\caption{RQ2 Monte Carlo uncertainty estimated from 120 held-out datasets per design.}
\label{tab:stage2_uncertainty}
\setlength{\tabcolsep}{4pt}
\begin{tabular}{lcccc}
\toprule
Design & Mean KL & Dataset SD & 95\% CI & MAP \\
\midrule
\texttt{d2b2} & 0.000 & 0.000 & [0.000, 0.000] & 0.167 \\
\texttt{d3b2} & 0.974 & 0.297 & [0.921, 1.028] & 0.650 \\
\texttt{d4b2} & 1.316 & 0.293 & [1.263, 1.370] & 0.800 \\
\texttt{d5b2} & \textbf{1.401} & 0.283 & [1.350, 1.453] & \textbf{0.883} \\
\bottomrule
\end{tabular}
\end{table}

\paragraph{RQ3 seed variability:}

RQ3 uncertainty was evaluated using a fixed reference configuration ($\alpha=0.75$, $\lambda=0.20$, hierarchical raw-state features, 960 training datasets) trained with five independent seeds. Table~\ref{tab:stage3_seed_variability} summarizes the KL-head predictions. The KL head ranked \texttt{d5b2} above \texttt{d4b2} in all five runs, with a mean difference of $0.126$ (SD $=0.014$; 95\% CI $[0.109,0.144]$). Posterior-derived KL produced the same ordering but with a smaller separation ($0.035$, 95\% CI $[0.017,0.053]$), reflecting imperfect posterior calibration. In contrast, posterior recovery did not clearly distinguish \texttt{d4b2} and \texttt{d5b2}. Differences in posterior probability assigned to the true latent state ($-0.012$, 95\% CI $[-0.028,0.004]$) and joint MAP accuracy (95\% CI $[-0.015,0.015]$) were statistically indistinguishable. Thus, stable design ranking is achieved even when posterior recovery remains uncertain. Overall, the reference surrogate achieved joint MAP accuracy of $0.407\pm0.008$, KL-head RMSE of $0.235\pm0.003$, perfect agreement with the exact RQ2 ranking ($\rho=1.0$), and zero design-selection regret across all five runs. These results indicate that the learned KL head provides more stable design evaluation than posterior-derived information gain and that design ranking is more robust than exact posterior recovery.

\begin{table}[t]
\centering
\caption{RQ3 uncertainty across five training seeds.}
\label{tab:stage3_seed_variability}
\setlength{\tabcolsep}{4pt}
\begin{tabular}{lccc}
\toprule
Design & Mean KL-head & Seed SD & 95\% CI \\
\midrule
\texttt{d2b2} & 0.034 & 0.007 & [0.025, 0.043] \\
\texttt{d3b2} & 1.000 & 0.022 & [0.973, 1.027] \\
\texttt{d4b2} & 1.293 & 0.009 & [1.281, 1.304] \\
\texttt{d5b2} & \textbf{1.419} & 0.011 & [1.405, 1.433] \\
\bottomrule
\end{tabular}
\end{table}

\subsection{Calibration of Predicted Information Gain}

\begin{figure*}
    \centering \includegraphics[width=0.88\textwidth,keepaspectratio]{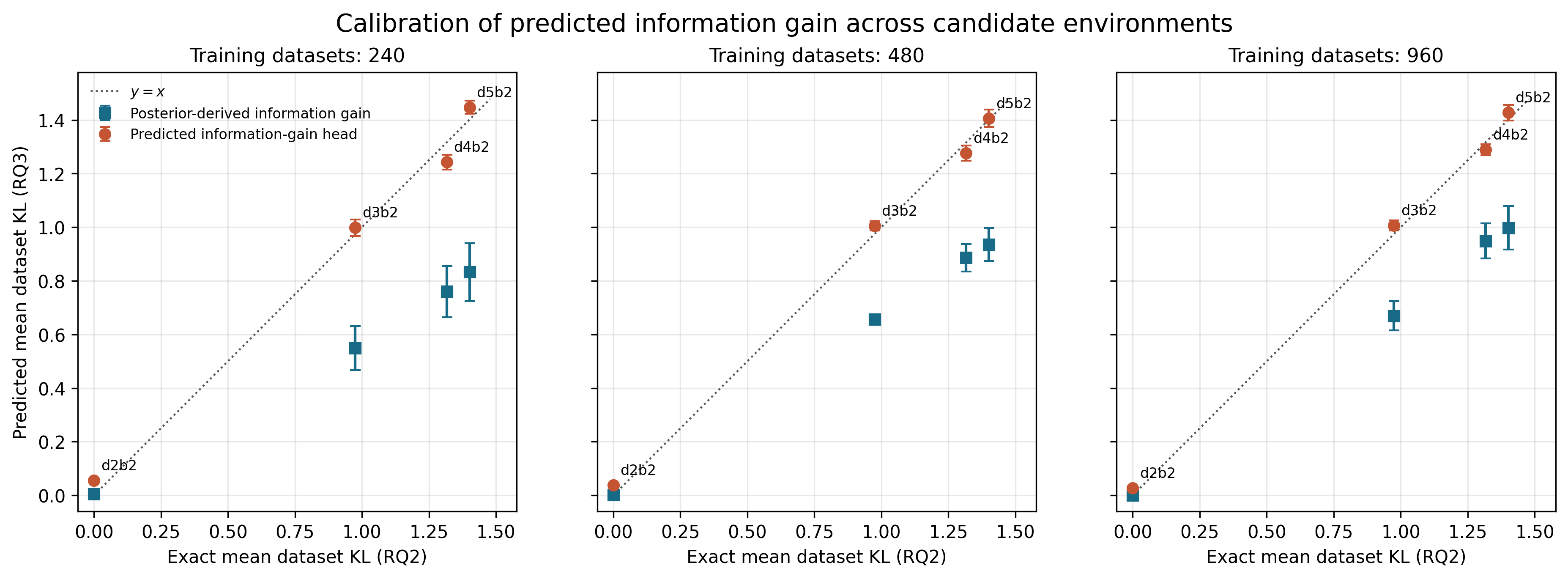}
    \caption{Calibration of predicted posterior-to-prior KL across candidate environments for joint inference of \((H,K)\) under different training-set sizes. The x-axis shows the exact MC-BED mean dataset KL and the y-axis the corresponding surrogate prediction. Blue squares denote KL computed from the amortized posterior, orange circles the auxiliary KL prediction head, and the dashed line indicates perfect calibration (\(y=x\))}
    \label{fig:kl_calibration}
\end{figure*}

Figure~\ref{fig:kl_calibration} compares the RQ3 predicted mean dataset-level information gain with the exact RQ2 MC-BED estimates across candidate environments. The horizontal axis shows the exact mean posterior-to-prior KL, the vertical axis the surrogate prediction, and the diagonal denotes perfect calibration. We compare the posterior-derived information gain, computed from the amortized posterior, with the auxiliary information-gain head trained to regress the exact MC-BED utility. The information-gain head is well calibrated across all training sizes (240, 480, and 960 datasets), closely matching the identity line and recovering the correct ordering
$\texttt{d5b2}>\texttt{d4b2}>\texttt{d3b2}>\texttt{d2b2}$. In contrast, the posterior-derived estimate consistently underestimates the exact KL, particularly for the more informative environments, although it largely preserves the correct ranking. This underestimation arises because the amortized posterior remains less concentrated than the exact posterior, reducing the induced posterior-to-prior KL. Overall, the results show that learning the scalar MC-BED utility is easier than recovering the full posterior. Direct KL supervision therefore produces well-calibrated design values even when posterior recovery remains imperfect, consistent with the ablation results.

\section{Computational Complexity}

Let $N$ denote the number of training datasets, $R$ the number of trajectories per dataset, $L$ the mean trajectory length, $F$ the input feature dimension, $U$ the hidden dimension, and $E$ the number of training epochs. A single GRU update requires $O(FU+U^2)$ operations per decision step, yielding a cost of $O(RL(FU+U^2))$ to encode one trajectory dataset and a total training complexity of $O(ENRL(FU+U^2))$. After training, evaluating a candidate dataset requires only a single forward pass with complexity $O(RL(FU+U^2))$.

In contrast, exact MC-BED performs recursive planning and exact posterior inference for every candidate environment, latent parameter setting, and Monte Carlo dataset, giving a complexity proportional to $O(|\Xi|\,|\Theta|\,M\,C_{\mathrm{exact}})$, where $|\Xi|$ is the number of candidate environments, $|\Theta|$ the number of latent parameter settings, $M$ the number of Monte Carlo datasets, and $C_{\mathrm{exact}}$ the cost of exact posterior inference for a single dataset. Amortization therefore replaces repeated exact inference with a fixed-cost neural forward pass during design evaluation.

\section{Additional Ablations}

\paragraph{Planning-Horizon Inference}
Unless otherwise stated, planning-horizon ablations use five binary environments, $H\in\{1,2,3\}$, fixed $K=2$, five trajectories per dataset, and five random seeds.

\paragraph{Loss weights}
We evaluate $\alpha\in\{0,0.25,0.5,0.75,1\}$ and
$\lambda\in\{0,0.1,0.2,0.5,1\}$. Planning-horizon MAP remains stable
($0.44$--$0.46$) across the grid. Without KL supervision ($\lambda=0$), posterior-derived ranking typically selects \texttt{d4b2}, incurring regret $0.128$. For $\lambda\ge0.2$, the KL head recovers the optimal design (\texttt{d6b2}) with Spearman $\rho=0.98$--$1.00$ and zero regret. Varying $\alpha$ has only a minor effect.

\paragraph{Representation}
Raw-state features achieve the highest $H$-MAP, and yields the strongest posterior ranking. All representations recover the optimal design through the supervised KL head.

\begin{table}[t]
\centering
\caption{Representation ablation for planning-horizon inference (mean over five seeds)}
\label{tab:representation_ablation_h}
\setlength{\tabcolsep}{4pt}
\begin{tabular}{lcccc}
\toprule
\textbf{Representation} &
\textbf{$H$-MAP} &
\shortstack{\textbf{Posterior}\\\textbf{$\rho$}} &
\shortstack{\textbf{KL-head}\\\textbf{$\rho$}} &
\textbf{Regret} \\
\midrule
Basic                  & 0.422 & 0.68 & 1.00 & 0 \\
Path-aware             & 0.443 & 0.66 & 1.00 & 0 \\
Raw-state              & \textbf{0.446} & \textbf{0.82} & 0.98 & 0 \\
\bottomrule
\end{tabular}
\end{table}

\paragraph{Training size}
Increasing the training set improves KL calibration (lower RMSE) but has little effect on $H$-MAP. Since ranking is evaluated over only five candidate designs, rank correlation should be interpreted together with RMSE and posterior accuracy.

\begin{table}[t]
\centering
\caption{Training-size ablation for planning-horizon inference}
\label{tab:training_size_ablation_h}
\setlength{\tabcolsep}{4pt}
\begin{tabular}{rcccc}
\toprule
\textbf{Train Size} &
\textbf{$H$-MAP} &
\shortstack{\textbf{Posterior}\\\textbf{$\rho$}} &
\shortstack{\textbf{KL-head}\\\textbf{$\rho$}} &
\shortstack{\textbf{KL-head}\\\textbf{RMSE}} \\
\midrule
150 & 0.442 & 1.00 & 1.00 & 0.077 \\
375 & 0.442 & 0.78 & 1.00 & 0.055 \\
750 & 0.448 & 0.68 & 1.00 & \textbf{0.041} \\
\bottomrule
\end{tabular}
\end{table}

\paragraph{Joint Inference}

Joint-inference ablations use four binary environments with
$H\in\{1,2,3\}$, $K\in\{2,5\}$, and ten trajectories per dataset. Joint MAP
remains between $0.38$ and $0.42$, well above the six-class chance level
($0.167$). Most configurations recover the correct design ordering, although
the ranking problem is relatively coarse because informativeness increases
approximately with tree depth. Path-aware features achieve the highest joint MAP. Increasing the training set
from 240 to 960 examples improves joint MAP from $0.338$ to $0.406$, while the
optimal design is recovered at all training sizes.

\paragraph{Summary of Ablation}
Across both settings, direct KL supervision is the primary driver of accurate design ranking. Richer representations and larger training sets mainly improve posterior recovery and KL calibration, whereas design ranking saturates much earlier.

\begin{table}
\centering
\caption{Heuristic environment-selection baselines}
\label{tab:baseline_heuristics}
\setlength{\tabcolsep}{4pt}
\begin{tabular}{lccc}
\toprule
\textbf{Method} &
\textbf{Spearman $\rho$} &
\textbf{Best} &
\textbf{Top-2} \\
\midrule
Random                & 0.002   & --             & -- \\
Largest tree          & 1.000   & \texttt{d5b2} & 1.00 \\
Cheapest tree         & -1.000  & \texttt{d2b2} & 0.00 \\
Minimum JS            & 1.000   & \texttt{d5b2} & 1.00 \\
Mean JS               & 0.800   & \texttt{d4b2} & 1.00 \\
Entropy reduction     & 0.800   & \texttt{d4b2} & 1.00 \\
Behavioural diversity & 0.800   & \texttt{d4b2} & 1.00 \\
Mean clicks           & 1.000   & \texttt{d5b2} & 1.00 \\
\bottomrule
\end{tabular}
\end{table}

\begin{table}
\centering
\caption{Supervised cognitive-state classification baselines}
\label{tab:classifier_baselines}
\setlength{\tabcolsep}{4pt}
\begin{tabular}{lcccc}
\toprule
\textbf{Method} &
\textbf{Train} &
\textbf{Validation} &
\textbf{Test} &
\textbf{Rank $\rho$} \\
\midrule
Logistic regression & 0.374 & 0.315 & 0.329 & 0.80 \\
Random forest       & 0.755 & 0.327 & 0.315 & 0.80 \\
Gradient boosting   & 0.757 & 0.292 & 0.283 & 1.00 \\
\bottomrule
\end{tabular}
\end{table}

\begin{table}
\centering
\caption{Direct regression baselines for MC-BED utility prediction}
\label{tab:regression_baselines}
\setlength{\tabcolsep}{3pt}
\begin{tabular}{lccccc}
\toprule
\textbf{Method} &
\textbf{RMSE} &
\textbf{Dataset $\rho$} &
\textbf{Dataset $\tau$} &
\textbf{Design $\rho$} &
\textbf{Best} \\
\midrule
Linear \\ Regression & 0.254 & 0.783 & 0.590 & 1.000 & \texttt{d5b2} \\
Random \\ Forest     & \textbf{0.250} & \textbf{0.802} & \textbf{0.624} & 1.000 & \texttt{d5b2} \\
Gradient \\ Boosting & 0.253 & 0.783 & 0.596 & 1.000 & \texttt{d5b2} \\
\bottomrule
\end{tabular}
\end{table}

\section{Baseline Methods}
\label{sec:baselines}

We compare ABED with three baseline families: (i) design-selection heuristics, (ii) supervised cognitive-state classifiers, and (iii) direct KL regressors. Heuristics rank environments without inference, classifiers estimate latent cognitive states and derive design scores from the predicted posterior, while regressors directly predict the MC-BED utility.

\paragraph{Design-Selection Heuristics}

For the joint-inference benchmark, the exact MC-BED ranking is
\texttt{d5b2}$>$\texttt{d4b2}$>$\texttt{d3b2}$>$\texttt{d2b2}.
Table~\ref{tab:baseline_heuristics} compares structural, behavioural, and posterior-based heuristics against this reference. Largest-tree and mean-click heuristics recover the optimum because informativeness is approximately monotonic with tree depth in this restricted binary benchmark. This relationship weakens in the expanded binary--ternary benchmark, where tree size alone no longer predicts the MC-BED ranking.

\paragraph{Cognitive-State Classifiers}

We train logistic regression, random forest, and gradient boosting to predict the joint latent state $(H,K)$ from behavioural summaries in Table \ref{tab:classifier_baselines}. Environment scores are obtained from the KL divergence between the predicted class distribution and the prior. Classification accuracy and design ranking are only weakly coupled. Gradient boosting recovers the correct ranking despite modest test accuracy, whereas random forest achieves higher training accuracy but selects a suboptimal design.

\paragraph{Direct KL Regression}

We also train linear regression, random forest, and gradient boosting to predict the exact dataset-level KL divergence $\mathrm{KL}(p(\theta\mid\mathcal D,\xi)\|p(\theta))$ from environment descriptors and behavioural summaries. Dataset predictions are averaged to obtain environment scores as shown in Table \ref{tab:regression_baselines}. All regressors recover the correct design ordering, with random forest giving the best dataset-level KL prediction. Unlike ABED, however, they estimate only a scalar utility and cannot perform posterior inference or posterior-guided adaptive experiment selection. Posterior distillation and KL-aligned objectives are treated as ablations of ABED rather than external baselines.

\section{Limitations and Future Work}

The current evaluation is restricted to symmetric tree-structured Mouselab-MDP environments. Although the expanded binary--ternary benchmark demonstrates that topology alone does not determine environment value, richer design spaces involving asymmetric trees, heterogeneous reward priors, varying click costs, and non-tree cognitive tasks remain important directions for future work. Larger environment families and additional cognitive tasks are needed to evaluate generalization beyond tree-structured decision problems. Furthermore, the present implementation relies on supervised KL targets obtained from MC-BED. Developing surrogate objectives that require fewer exact design evaluations remains an important direction for scaling amortized Bayesian experimental design.

\end{document}